\documentclass[letterpaper, 10 pt, journal]{ieeeconf} % Comment this line out if you need a4paper

\IEEEoverridecommandlockouts               % This command is only needed if 
\usepackage{graphics} % for pdf, bitmapped graphics files
\usepackage{epsfig} % for postscript graphics files
\usepackage{amsmath} % assumes amsmath package installed
\usepackage{amssymb} % assumes amsmath package installed
\usepackage{tabularx}
\usepackage{multirow}
\usepackage{color}
\usepackage{url}
\usepackage{breqn}
\newcommand{\junk}[1]{}
\usepackage{algorithm2e}
\usepackage[export]{adjustbox}
\usepackage{tabulary,booktabs}
\usepackage{comment}
\usepackage{bm}
\usepackage{subfigure}
\usepackage{hyperref}
\usepackage{soul}

\newcommand{\norm}[1]{\left\lVert#1\right\rVert}
\newcommand*{\eg}{e.g.\@\xspace}
\newcommand*{\ie}{i.e.\@\xspace}

\newcommand{\figref}[1]{Fig.~\ref{#1}}
\newcommand{\secref}[1]{Section~\ref{#1}}
\newcommand{\tabref}[1]{Table~\ref{#1}}

\title{\LARGE \bf
TeLeMan: Teleoperation for Legged Robot Loco-Manipulation\\ using Wearable IMU-based Motion Capture
}

\author{Chengxu Zhou\textsuperscript{1}, Christopher Peers\textsuperscript{1}, Yuhui Wan\textsuperscript{1}, Robert Richardson\textsuperscript{1}, and Dimitrios Kanoulas\textsuperscript{2}% <-this % stops a space
\thanks{\textsuperscript{1}School of Mechanical Engineering, University of Leeds, Woodhouse Lane, Leeds LS2 9JT, UK. {\tt\small c.x.zhou@leeds.ac.uk}}%
\thanks{\textsuperscript{2}Department of Computer Science, University College London, Gower Street, London WC1E 6BT, UK. {\tt\small d.kanoulas@ucl.ac.uk}}%
\thanks{This work involved human subjects in its research. Approval of all ethical and experimental procedures and protocols was granted by the Business, Environment and Social Sciences Research Ethics Committee of the University of Leeds (Reference No. MEEC 21-009).} 
\thanks{This work was supported by the Dstl [grant number DSTLX1000156070], the EPSRC [grant numbers EP/R513258/1-2441459, EP/V026801/2], the AMPI [Innovate UK project number 84646] and the UKRI Future Leaders Fellowship [grant number MR/V025333/1].}% <-this % stops a space
}
\begin{document}

\maketitle
% \thispagestyle{empty}
% \pagestyle{empty}

%%%%%%%%%%%%%%%%%%%%%%%%%%%%%%%%%%%%%%%%%%%%%%%%%%%%%%%%%%%%%%%%%%%%%%%%%%%%%%%%
\begin{abstract}
Human life is invaluable. When dangerous or life-threatening tasks need to be completed, robotic platforms could be ideal in replacing human operators. Such a task that we focus on in this work is the Explosive Ordnance Disposal. Robot telepresence has the potential to provide safety solutions, given that mobile robots have shown robust capabilities when operating in several environments. However, autonomy may be challenging and risky at this stage, compared to human operation. Teleoperation could be a compromise between full robot autonomy and human presence. In this paper, we present a relatively cheap solution for telepresence and robot teleoperation, to assist with Explosive Ordnance Disposal, using a legged manipulator (\ie, a legged quadruped robot, embedded with a manipulator and RGB-D sensing). We propose a novel system integration for the non-trivial problem of quadruped manipulator whole-body control. Our system is based on a wearable IMU-based motion capture system that is used for teleoperation and a VR headset for visual telepresence. We experimentally validate our method in real-world, for loco-manipulation tasks that require whole-body robot control and visual telepresence.
\end{abstract}

%%%%%%%%%%%%%%%%%%%%%%%%%%%%%%%%%%%%%%%%%%%%%%%%%%%%%%%%%%%%%%%%%%%%%%%%%%%%%%%%
\section{Introduction}
In the area of defence and security, there are often tasks that may be too risky for a human to attempt to complete. For instance, in the UK, there are more than $2,500$ yearly operations for Explosive Ordnance Disposal (EOD). The human expertise involved in completing such tasks is at an impressive level. Therefore, due to human safety and well-being purposes, a robotic alternative is sought by defence departments around the world.

Robotic platforms, that are able to navigate in challenging environments and manipulate objects, would be ideal for operating in the aforementioned demanding tasks. Legged quadrupedal robots have become cheaper, more robust in unstructured environments (e.g., rough terrains and stairs), with high computational capabilities (e.g., GPUs), and a wide range of on-board sensors (e.g., IMUs, cameras, force/torque sensors). Very recently, researchers started embedding robotic arms on the top of quadruped robots to enhance them with manipulation capabilities. For instance, such robots include Spot (by Boston Dynamics), ANYmal (by ANYbotics), HyQ and CENTAURO (by IIT). Such a platform--a legged quadruped manipulator--is ideal in completing loco-manipulation tasks.

While a big part of the robotics community has focused on autonomous scene understanding and action performance, achieving human-level skill quality is not trivial. Robust high-level decision-making for several complex tasks, is still under heavy research. Thus, controlling the high-level actions of robots via teleoperation and telexistence is a good compromise on the path towards full autonomy. Even though robot teleoperation and telexistence is not a new topic of study~\cite{Mason2012}, it has attracted the interest of the research community again, given that only recently, the computational units, sensors, and wireless connection capabilities have been developed to such a level that might allow for real-time robust performance, up to an extent.

\begin{figure}[t!]
  \centering
  \includegraphics[width=\columnwidth]{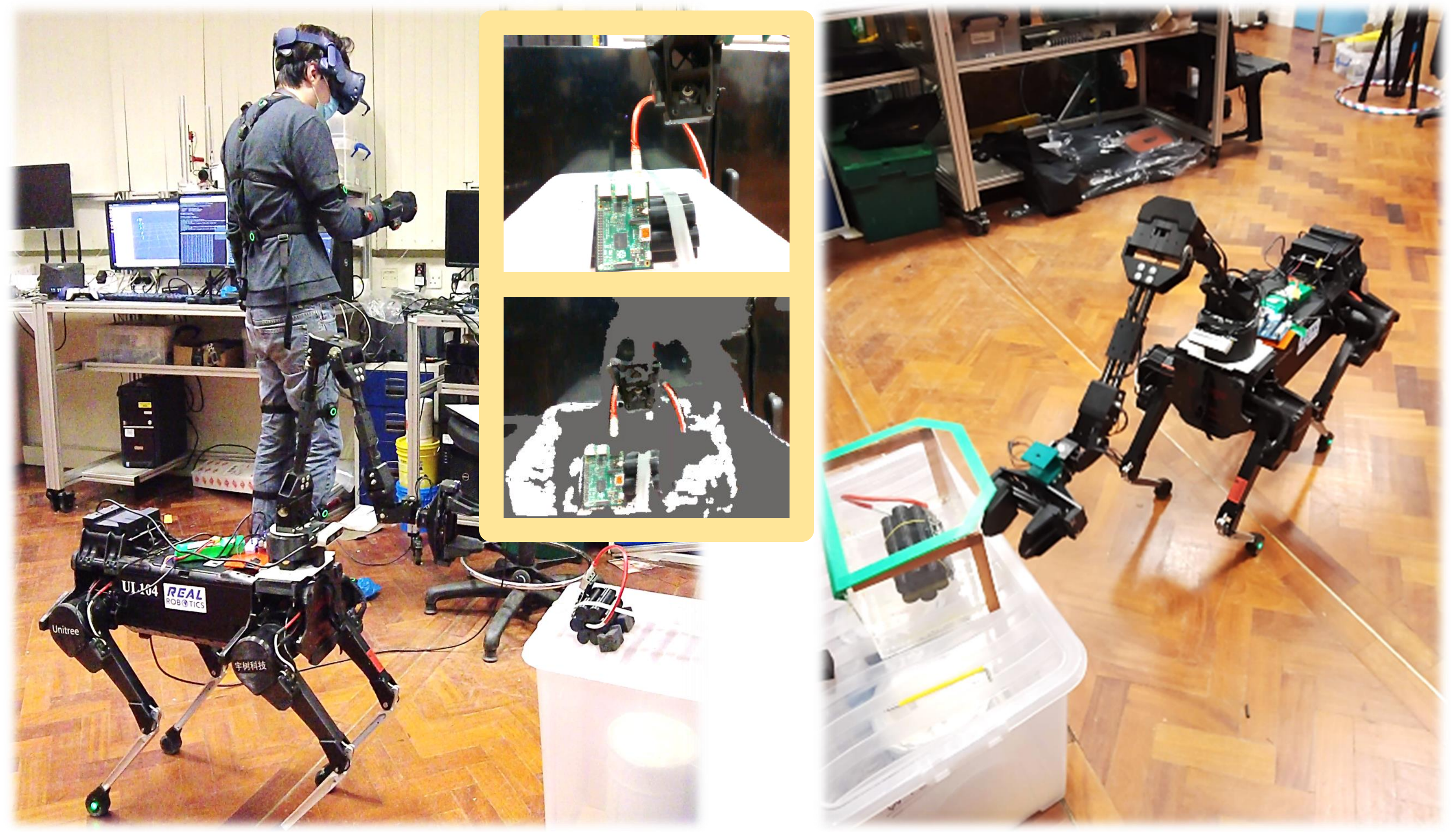}
  \caption{Teleoperative legged manipulator system: a teleoperator with VR headset and wearable IMU-based motion capture, controlling locomotion and manipulation.}
  \label{fig:system-intro}
\end{figure}

In this paper, we firstly introduce hardware adjustments to assemble a low-cost, but yet competent, quadrupedal manipulator (\figref{fig:system-intro}). It includes a quadruped robot (Unitree Laikago), a modified 5 Degrees-of-Freedom (DoF) robotic arm (ViperX 300), and an RGB-D camera (Intel Realsense). Secondly, we introduce a unified system for whole-body loco-manipulation control, using a wearable IMU-based motion capture system for teleoperation (Perception Neuron Motion Capture), via $5$ GHz Wi-Fi $6$, for real-time communication. Based on this system, a teleoperator can control simultaneously the robot locomotion and manipulation in an intuitive manner, while receiving visual feedback from the depth camera, via a Virtual Reality (VR) headset (HTC Vive Pro). The goal is to allow easy real-time robot teleoperation and telexistence for risky defence and security tasks, such as the EOD. Finally, real-world experiments for EOD tasks were used to evaluate the full system.

\begin{figure*}
  \centering
  \includegraphics[width=0.93\textwidth]{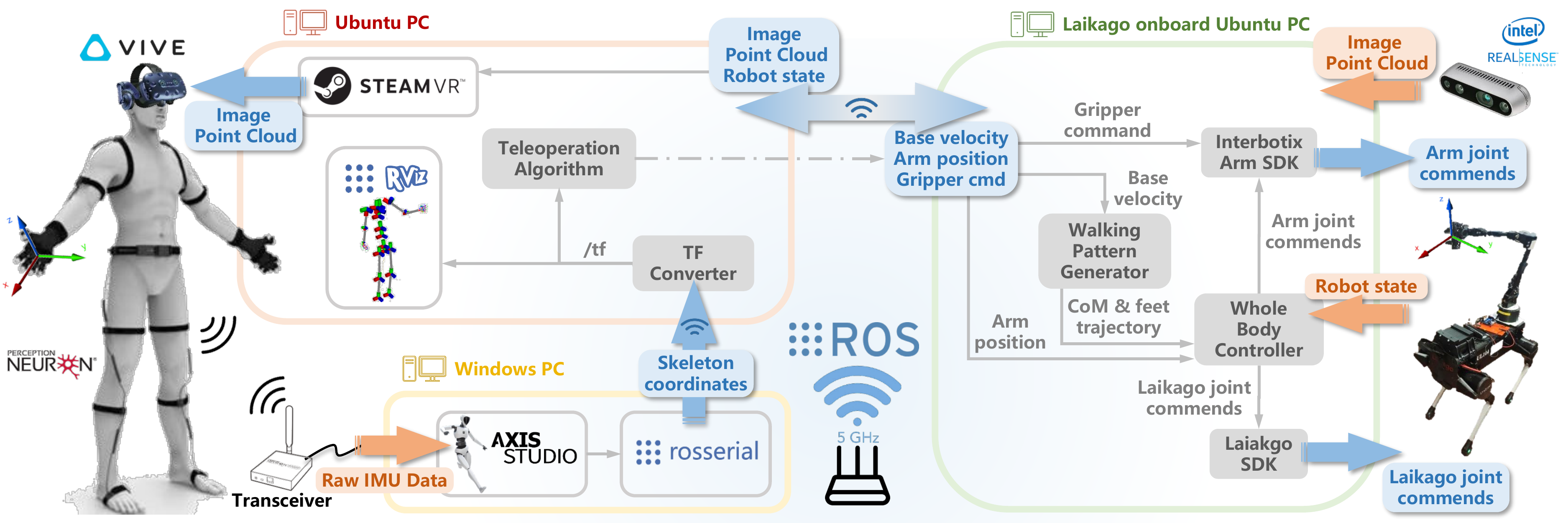}
  \caption{System overview: the teleoperator (left) controlling the robot base and arm (right), via wearable IMU-based motion capture, while getting feedback from the robot's RGB-D camera via the VR headset.}
  \label{fig:overview}
\end{figure*}

Overall, the main contributions of this paper are as follows:
\begin{enumerate}
    \item introducing a novel hardware development for an inexpensive quadrupedal manipulator, by also reconfiguring the robotic arm into a lightweight one, saving payload effort for the robot,
    \item introducing a new unified teleoperation system for loco-manipulation, based on a wearable IMU-based motion capture system; a system that has not been used in the research quadrupedal legged robotics community before, and
    \item demonstrating via real-world experiments the superiority of the telexistence and teleoperation capability to complete Explosive Ordnance Disposal (EOD) tasks, compared to traditional gamepad methods.
\end{enumerate}

The remainder of the paper is structured as follows. In \secref{Sec:rw}, we present the related work in robot teleoperation and telexistence, followed by the hardware system description in \secref{Sec:hw}. In \secref{Sec:sw}, we present the whole-body teleoperation system, while in \secref{Sec:exp} we present the experimental validation of the system on the real robot. Finally, we discuss conclusions and future work in \secref{Sec:concl}.

%%%%%%%%%%%%%%%%%%%%%%%%%%%%%%%%%%%%%%%%%%%%%%%%%%%%%%%%%%%%%%%%%%%%%%%%%%%%%%%%
\section{Related Work}\label{Sec:rw}
Joystick control was the standard for legged/wheeled robot teleoperation~\cite{Chestnutt2006, Klamt2018} until recently. Even though joysticks are ideal when a teleoperator needs to control the position or velocity of a single part of a robot, e.g., its body on flat ground, it becomes challenging and non-trivial when: 1) more complex movements are required, e.g., the $6$DoF motion of a manipulator's end-effector (which can be resolved with motion capture hand controllers~\cite{Zhang2018}), or 2) multiple parts of a robot need to be controlled simultaneously, e.g., the position/velocity of its body and a different position/velocity of its end-effector. We have extensively noticed these disadvantages on legged manipulators. For instance, two examples include our work on the IIT-Centauro animaloid robot~\cite{Parnell2018} and the work on ETH-ANYmal legged manipulation robot~\cite{anymalwitharm}. For this reason, the research community has focused on different devices and sensors for teleoperating complex legged manipulators.

A common alternative method to teleoperate complex multi-DoF robots is to create a wearable exoskeleton twin for the teleoperator. Even though this is an expensive alternative, it has been only successful with manipulation tasks~\cite{Klamt2020}. While usually navigation is controlled firstly with a joystick, disengaging the two tasks makes the system too complex and challenging to be used. On the other side, locomotion has been handled in the past by using treadmills~\cite{Elobaid2018}, which is a rather expensive solution to the problem.

Several researchers are using motion capture systems to control robots, mainly again for manipulation tasks only. For instance, in~\cite{Deshpande2018}, motion capture for the position of the arm, along with a hand/fingers haptic device, was used to teleoperate a quadrupedal manipulator. Although, again locomotion was not demonstrated via these devices, thus the robot navigation was controlled before the manipulation task. Moreover, motion capture systems have the disadvantage that they need to be installed in a room or space, while ideally a human teleoperator should be able to move anywhere to operate the robot. 

Following a new trend in robotics, originating from the animation industry, our work is focused on using an IMU-based wearable suit to teleoperate the robot. Previous works show that IMU-embedded wearable motion capture devices can be used reliably enough to generate high and low-level control for robots with a high degree of accuracy, furthering its potential in complex manipulation tasks. One example is presented in~\cite{imumocap1}, where the high-level walking motion is used to control a bipedal robot, with the teleoperator using foot positioning to determine the robot actions. Another example, presented in~\cite{imumocap2}, demonstrates the use of a gesture-based control scheme, via an IMU embedded device on the hand and arm, to control a wheeled robot. Likewise, wearable IMU-based motion capture systems are also possible to reliably control $7$DoF robotic arms~\cite{imumocap3}. Wearable IMU-based motion capture devices have also been demonstrated to be accurate and robust in domestic scenarios, e.g., for a dementia-care robot~\cite{imumocap4}. In~\cite{eda}, the TALOS bipedal robot's whole-body is controlled as a fixed-base system (i.e., without locomotion), using an Xsens 3D motion tracking suit. In the system realisation, the hands and head tracking is used with an absolute Cartesian approach to achieve robotic teleoperation. The teleoperator's motion capture suit data is completely overlaid on the humanoid robot. An additional human model is used as a method to format the suit data to fit the robot. Our work, is the first we are aware of, that uses an IMU-suit to control simultaneously the navigation and manipulation of a quadrupedal legged manipulator.

Regarding telexistence, the majority of the current methods include VR headsets~\cite{Elobaid2019}, and haptic devices~\cite{Tachi2013}, to give enough visual, audio, or haptic feedback to the teleoperator, in order to control the robot efficiently and accurately. We use similar vision-based VR-based methods in this work to provide the teleoperator with telexistence capabilities, leaving haptic feedback for future work.

%%%%%%%%%%%%%%%%%%%%%%%%%%%%%%%%%%%%%%%%%%%%%%%%%%%%%%%%%%%%%%%%%%%%%%%%%%%%%%%%

\section{Hardware System Description}\label{Sec:hw}
In this section, we provide a description of all hardware that was used for robot teleoperation and telexistence, as well as its modifications. In \figref{fig:overview}, an overall structure of the system is visualised, including: 1) an integrated robotic platform, consisting of a quadruped robot, a robotic arm with a gripper, and an RGB-D sensor, and 2) a human teleoperation system, consisting of a VR headset and an IMU-based body motion capture system.

The robot hardware can be separated into two parts. The first part is the robot base, which consists of the Unitree Laikago quadruped robot, with $5$ kg of payload. The robot base is powered by an on-board Intel NUC i5 computing system. On top of the quadruped robot base is the second part, which is a ViperX $300$ Robot Arm and a visual Intel RealSense RGB-D camera. The robotic arm was redesigned in order to be mounted on top of the quadruped robot. Also, the RGB-D camera is able to rotate with the arm around the yaw-axis. A full simulated model of the integrated robot parts was generated within the Robotic Operating System (ROS) and Gazebo, to test all its capabilities before experimenting on the real robot. 

\begin{figure*}
 \centering
 \includegraphics[width=0.93\textwidth]{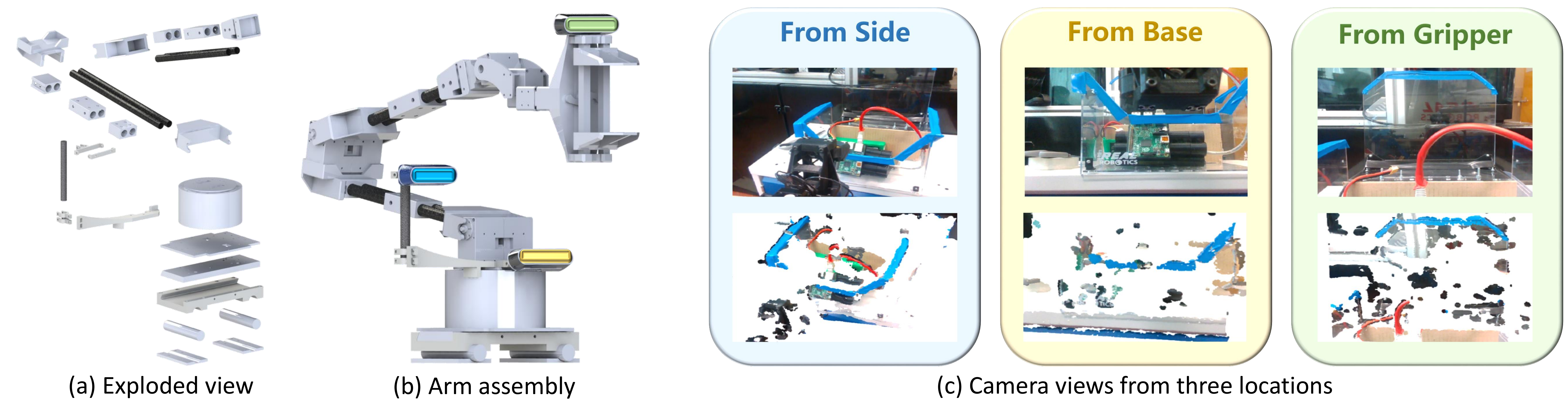}
 \caption{(a) Exploded view of all redesigned components in the robotic arm, (b) full assembly of the robotic arm with three possible camera mounting positions coloured and (c) the corresponding RGB and depth camera image from each mounting location on the robotic arm.}
 \label{fig:teleman_arm}
\end{figure*}

For teleoperation, Noitom's Perception Neuron Studio, a wearable motion capture system based on inertia sensing, was used for precisely estimation of human body postures. The system consists of $16$ IMU sensors, which are attached to each link of a human body, and a pair of Studio Motion Capture Gloves for the finger estimation. This system provides accurate estimation of the human skeleton's positions and orientations in world coordinates, which include $19$ body segments and $40$ hand segments, at $100$ Hz with a minimum resolution of $0.02$ degrees. Calibration is needed for every new user which only takes about $30$ s.

For the VR system, a HTC VIVE Pro headset was used, which visualises the robot model, the human teleoperator as a transformation tree, as well as the point clouds and RGB images from the robot's Intel RealSense camera. Last, to be able to operate in a real-time sensitive environment, a $5$~GHz Wi-Fi router was used for the communication between all devices.

\subsection{Hardware Redesign}\label{sec:hwredesign}
One of the biggest constraints of using a quadruped robot base platform is its payload limit. A heavy payload not only increases the workload on the motors, but also increases the height of the center of mass. Therefore, it has a considerable impact on the robot's stability and maneuverability. To minimise this impact, various methods were utilised to reduce extra loads placed on the robot. The heaviest object in the design was the $5$ DoF ViperX $300$ robotic arm. In order to reduce the weight of the arm, while maintaining its capability, several parts have been redesigned. This involves, firstly, replacing the aluminum box-section links of the arm with carbon fibre rods. Secondly, redesigning the structure of the shoulder joints to maintain the robotic arm's capability with the new carbon fibre rods. Given the aforementioned changes, we managed to reduce the overall weight of the arm from $4.1$ kg to $2.5$ kg, a $39\%$ weight reduction. With the newly designed arm, the Laikago quadruped robot base can maintain strong stability and maneuverability, while still having an allowance for extra manipulation workload on the arm. This robot arm was mounted onto the Laikago's rear parallel carbon fiber rods, using a 3D-printed easily interchangeable slide-in mounting system. A diagram of the redesigned robot arm and the 3D-printed Laikago mounting system is shown fully rendered in \figref{fig:teleman_arm}.

An RGB-D Intel RealSense camera was used as visual feedback. This allows the operator to achieve telexistence via the RGB image and coloured depth-based point cloud representation of the world. In particular, the RGB images provided are essential for both locomotion and manipulation teleoperation tasks, however, the addition of the point clouds enables manipulation tasks that require a more accurate measure of distance. Point clouds allow the operator to visually confirm and pinpoint the three-dimensional locations of specific details in the environment and could provide great assistance while dealing with complex tasks that require a heightened sense of target and surrounding environments, such as opening boxes or cutting wires. 
Various mounting positions for the RGB-D camera were tested during the experimental validation to locate the most suitable position and angle for teleoperation. We mounted the RGB-D camera onto a part of the robot arm that allows the camera to maintain its perspective, regardless of the robot arm's yaw angle. We tested three mounting positions: 1) on the base of the robot arm, in the center pitched slightly upwards, 2) on the wrist joint of the robot arm behind the end-effector, and 3) to the side of the arm at a set height rotated towards the arm slightly in the yaw axis. A visual representation of these locations along with their respective 3D-printed mounting parts is shown in \figref{fig:teleman_arm}.

During real-world experiments, we noticed that, when a manipulation task was aimed (e.g., when attempting to open a box cover or pull out a wire), it was difficult for the operator to perceive whether the end-effector was properly positioned, due to the arm itself was obstructing the vision of the top/bottom end-effector gripper respectively, making it extremely difficult to determine the vertical location of end-effector. This issue was however solved with the side camera positioning. This camera location gave a ``third-person perspective'' to the operator. In this way, the operator can easily determine both the horizontal and vertical location of the end-effector. This camera position also allowed for all key objects that were manipulated to be in the frame along with the end-effector, without any obscuring problems. The quality of the point cloud however differs in each position. It was observed that, while viewing from the base and gripper locations, the quality of the point cloud began to decrease due to the camera being too close to the object, which causes some points to not be readable.

%%%%%%%%%%%%%%%%%%%%%%%%%%%%%%%%%%%%%%%%%%%%%%%%%%%%%%%%%%%%%%%%%%%%%%%%%%%%%%%%
\section{Software System Description}\label{Sec:sw}
Having introduced the hardware design that was developed for teleoperation and telexistence above, this section describes the software architecture of the system. The system can be split into three parts: robot motion generation, IMU-sensing for teleoperation with strategies, and the generation of the VR environment that enables telexistence.

\subsection{Gait Generation and Whole-Body Control}
Based on the virtual leg concept~\cite{sutherland1984}, by considering a pair of legs that move simultaneously as a single one, several quadrupedal robotic gaits, \eg, pacing, trotting, and bounding, can be realised by adopting the gait pattern generators developed for bipedal robots. Therefore, we have extended our previous walking algorithm~\cite{zhou2017jbe} to generate the quadrupedal trotting gait by considering the diagonal legs as two virtual legs. As shown in \figref{fig:overview}, the generated trajectories are sent to a whole-body controller for coordinated loco-manipulation of the developed legged manipulator.

The whole-body controller that considers the full dynamics of the developed legged manipulator is formulated as a quadratic programming problem:
\begin{equation}
    \min_{\bm{\mathcal{X}}} \quad \sum_{i=1}^{n} \omega_i \norm{ \bm{A}_{i}\bm{\mathcal{X}}-\bm{b}_{i}}^2, 
    \label{eq:opt_obj}
\end{equation}
where the sum of $n$ task's costs is minimised to obtain the optimal value of the target variable, $\bm{\mathcal{X}}=[\ddot{\bm{q}}, \bm{\lambda}]^T$, which consists of the generalised acceleration $\ddot{\bm{q}}$ and contact wrenches $\bm{\lambda}$. The $i^{\text{th}}$ task is defined by an objective matrix and vector, $\bm{A}_i$ and $\bm{b}_i$, and a weight $\omega_i$ that determines the soft priorities between tasks. Please refer to~\cite{You2016robio} for detailed formulations of individual tasks.

During the optimisation process, the following constraints are considered:
i) floating base dynamics $\bm{M}_{f}\ddot{\bm{q}} + \bm{h}_{f} = \bm{J}_{f}^{T}\bm{\lambda}$, 
ii) joint torque limits $\bm{M}_{a}\ddot{\bm{q}} + \bm{h}_{a} - \bm{J}_{a}^{T}\bm{\lambda} = \bm{{\tau}} \in [\bm{\tau}_{\text{min}},\bm{\tau}_{\text{max}}]$,
iii) non-slip contact constraints $\bm{J}\ddot{\bm{q}} + \dot{\bm{J}}\dot{\bm{q}} = \bm{0}$, and 
iv) contact force constraints to keep each contact force within a linearised friction cone.
% for each contact $\abs{f_x/f_z}\leq\mu,\abs{f_y/f_z}\leq\mu,f_z>0$,
% where 
$\bm{M}$ is the inertia matrix, $\bm{h}$ is the sum of Coriolis, centrifugal and gravitational terms, $\bm{J}$ is the contact Jacobian matrix corresponding to the contact wrenches $\bm{\lambda}$ at all contact points, the subscript $f$ represents the top six rows of a matrix for the floating base and $a$ corresponds to the actuated DoFs, $\bm{q} = [\bm{q}_f, \bm{q}_a]^{T}$ are the generalised coordinates which consists of the 6 DoF floating-base coordinates $\bm{q}_f$, and the actuated joints $\bm{q}_a$, $\bm{\tau}$ is a vector of joint torques.

The optimised result $\bm{\mathcal{X}}$ is then used to calculate the joint torque, joint position, and joint velocity references for lower-level control of the developed legged manipulator~\cite{You2016robio}.

\subsection{Teleoperation Strategies}\label{subsec:tele}

\begin{figure}
    \centering
    \subfigure[IMU-suit]{\label{fig:MotionInt}
    \includegraphics[height=39mm]{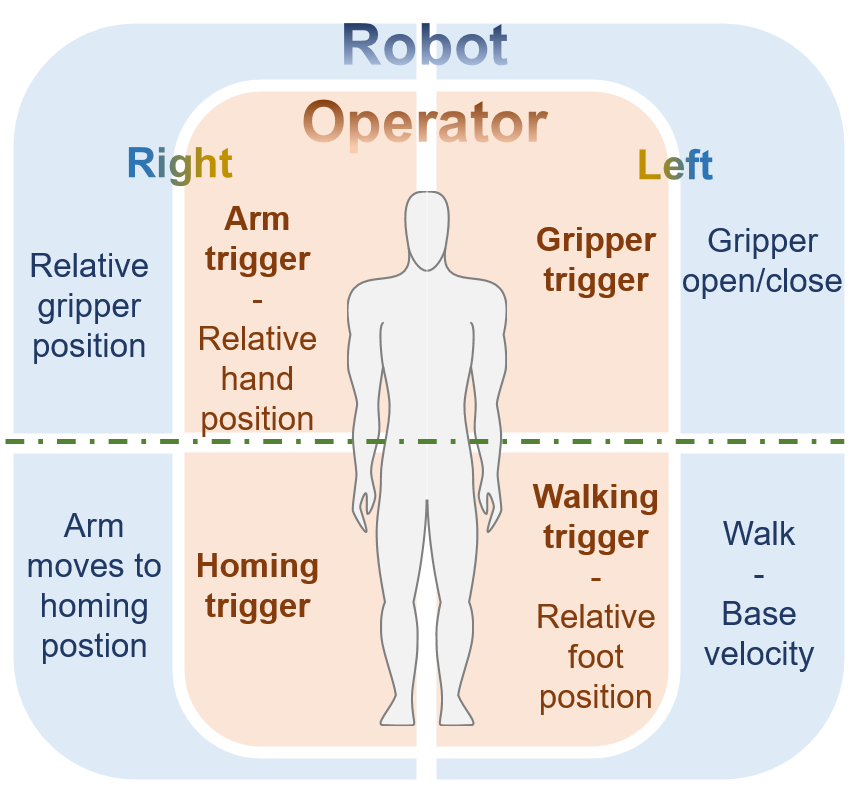}}
    
    \subfigure[Gamepad]{\label{fig:GamepadInt}
    \includegraphics[height=30mm]{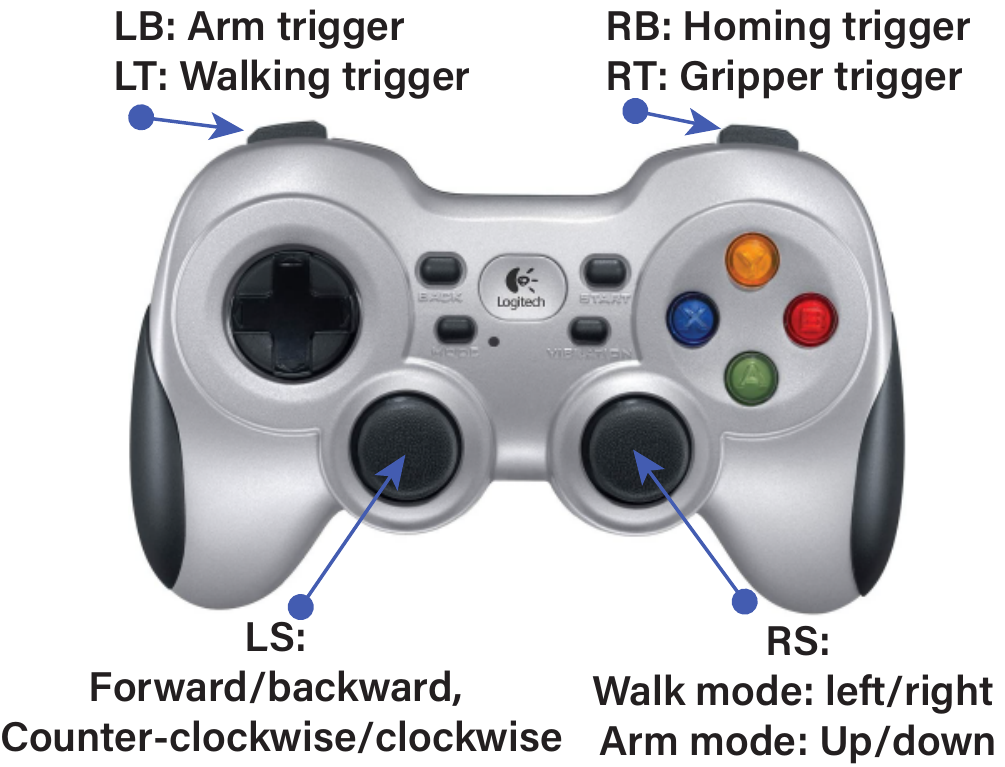}}
    
    \caption{Details of mapping from interfaces to trigger and argument strategies: (a) IMU-suit, (b) gamepad.}
    \label{fig:teleop}
\end{figure}

The human body motion is captured by a wearable motion capture system (Noitom Perception Neuron), as appears in \figref{fig:overview}-left. This provides stable and accurate human body segments pose estimations. The accompanied software (Noitom Axis Studio) can read full skeletal data, including fine finger movements. This data is then broadcasted to ROS through the \emph{rosserial} protocol. Since the human teleoperator and the robot are kinematically dissimilar, directly connecting them at a joint level is impractical. Thus, we have developed four strategies to intuitively teleoperate the legged manipulator in locomotion and manipulator tasks. As shown in \figref{fig:teleop}, the teleoperator uses hand closure as a trigger to send commands to the robot:
\begin{itemize}
  \item \emph{Gripper trigger} is activated when the left hand closes above the waist. The gripper on the manipulator will remain closed until this trigger is released.
  \item \emph{Walking trigger} is activated when the left hand closes below the waist. The teleoperator can then send base velocity references to the walking pattern generator by taking a step forward/backward, left/right, counter-clockwise/clockwise, or with a combination of directions. The step size or the rotation angle defines the magnitude of velocity that the robot will move at. The robot stops when this trigger is released.
  \item \emph{Arm trigger} is activated when the right hand closes above the waist. Then, the teleoperator's right arm movement will generate relative Cartesian arm position references to the whole body controller. The arm stops moving when this trigger is released.
  \item \emph{Homing trigger} is activated when the right hand closes below the waist. The robot joints will move to a predefined configuration.
\end{itemize}
The left and right-hand closure triggers can be combined, for instance, using left hand to enable \textit{walking trigger} and right hand for \textit{arm trigger} at the same time, thus performing simultaneous locomotion and manipulation control of the legged manipulator. 

Relative scaled pose relationship is used for connecting the teleoperator and robot movements, where at time \emph{t}, the \textit{arm} and \textit{walking triggers} are modelled as:
\begin{equation} \label{eq:tele}
    \begin{cases}
        \bm{a}_{rd}^{t}=\bm{a}_{rd}^{0}+\bm{\mu}(\bm{a}_{e}^{t} - \bm{a}_{e}^{0}) &\emph{arm trigger}\\
        \dot{\bm{w}}_{rd}^{t}=\dot{\bm{w}}_{rd}^{0}+\bm{\mu}(\bm{w}_{e}^{t} - \bm{w}_{e}^{0}) &\emph{walking trigger}
    \end{cases}
\end{equation}
where $\bm{a}=[x, z, \theta_{\text{roll}}, \theta_{\text{pitch}}, \theta_{\text{yaw}}]$, $\bm{w}=[x, y, \theta_{\text{yaw}}]$ define the displacements in sagittal ($x$), lateral ($y$) and vertical ($z$) directions and rotation about the front-to-back axis ($\theta_{\text{roll}}$), the side-to-side axis ($\theta_{\text{pitch}}$) and the vertical axis ($\theta_{\text{yaw}}$) for arm and base movement, respectively, subscripts \emph{e} refers to the teleoperator, \emph{r} to the robot, and \emph{d} to a desired value, superscript \emph{0} refers to the initial timing when the trigger is activated, $\bm{\mu}$ is used to scale the motions between the teleoperator and the robot. The VR glove's readings are used to detect the hand's closures for activating the triggers to teleoperate the robot.

This trigger-based relative motion strategy eliminates the long-term sensor drifting issue, especially for the yaw angle, which normally imposes negative impact on the controllers that rely on absolute yaw sensing.

\subsection{VR-based Integration for Telexistence and Teleoperation}
\label{subsec:vr}
In many real-world applications, a direct line of sight between the robot and the teleoperator may not always be possible. Some missions even require the teleoperator to keep a safe distance on-scene, for instance, in confined space, HAZMAT, and EOD tasks. Therefore, a remote way to monitor the robot movement and surrounding environment is necessary for these situations. Furthermore, since the wearable motion capture system uses human body motion as a control input signal from the teleoperator, it will not be ergonomic to look at a fixed display while the teleoperator’s body turns around. VR also introduces a full telexistence experience, which provides more information than traditional display and allows robust and efficient robot control. In the experiment, the teleoperator cannot perceive any aspect of the experiment except for the information displayed within the VR headset to maintain the teleoperative aspect of the experiment.

To generate the VR environment, the view of the robot's and teleoperator's state along with visual RGB-D feedback is generated and integrated into the teleoperator's view. First, the robot's RGB-D camera provides the teleoperator with a coloured 3D point-cloud and a 2D RGB image embedded above it (see \figref{fig:teleman_arm}-(c)). The RGB image generates a straightforward view of the surrounding environment, while the point cloud provided a more detailed structure. Notice that, the coloured images were compressed before communication to achieve higher transmission performance. The point clouds were cropped into the surrounding $2$ m distance only, to give a clearer and fast view of the setting. Various methods of filtering the point cloud were tested, such as downsampling (e.g., voxelisation) or statistical outlier removal. Although, it is hard to balance precision with minimal representation, objects such as wires, still occupy very few points in the cloud, and thus they cannot be expressed if such filtering takes place.

Secondly, the robot's model and the teleoperator's IMU-based skeleton view were visualised in the VR headset. In this way, the teleoperators were able to see their own body structure, the robot, and the visual environment around it for a full telexistence experience. The data between the robot and the VR headset were transmitted via $5$ GHz Wi-Fi to ensure bandwidth. Several experiments, e.g., VR-based manipulation, were also performed on a mobile manipulator at the early stage of development (UCL MPPL~\cite{Liu2020}).

%%%%%%%%%%%%%%%%%%%%%%%%%%%%%%%%%%%%%%%%%%%%%%%%%%%%%%%%%%%%%%%%%%%%%%%%%%%%%%%%
\section{Experimental Results}\label{Sec:exp}
A gamepad is another widely used control method for robots, and many quadruped manufacturers consider them as the default controller for commercial customers. In this research, we compare the performance difference between the wearable motion capture system and the gamepad. As a control, the teleoperation strategies are the same between the two controllers. They both have velocity control for locomotion and position control for manipulation. In detail, the trigger and bumper buttons on the gamepad are mapped to triggers in the teleoperation strategy and the joysticks are mapped to relative position and velocity. During the experiment, the user teleoperates a legged manipulator robot to complete four tasks with both types of controllers. The time took to finish each task is measured as their performance. 

\subsection{Experimental Design}
\begin{figure}
  \centering
  \includegraphics[width=0.9\columnwidth]{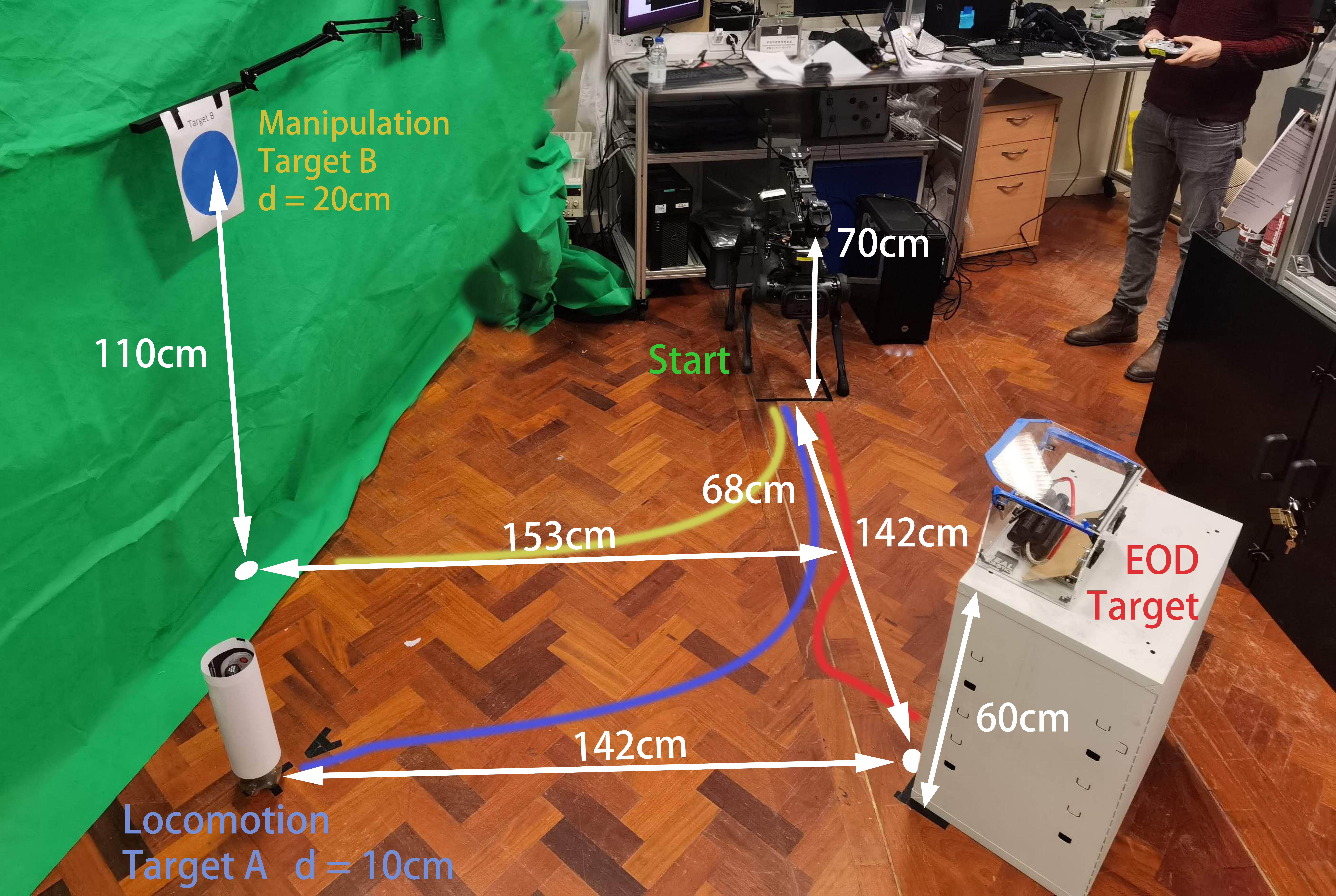}
  \vspace{-2mm}
  \caption{Detailed map of the experiment area with the locations of each target.}
  \label{fig:map}
\end{figure}

The experiment has been divided into four tasks, as shown in Fig.\ref{fig:map}. Users firstly perform basic manoeuvre of robot teleoperation to have a standard view with both controllers, then they try to accomplish a locomotion task, a manipulation task, a combined loco-manipulation task and an EOD task. The order of the first two tasks is randomised, half of the users perform the locomotion task first, and half perform the manipulation task first. After that, they perform the EOD task again in the same conditions, without direct sight through VR.
\begin{enumerate}
  \item The locomotion task requires the robot to walk from starting point to target A only. 
  \item The manipulation task starts with the robot standing below target B, and it requires moving the robot arm from the home pose to reach target B hanging in the air.
  \item The locomotion and manipulation task requires the robot to start from the starting point and walk towards target B, then use its arm to reach target B in the air. 
  \item The last task is the EOD task, a simulated real-world application to disassemble a bomb inside a box. In detail, the robot needs to walk from the starting point to the front of the bomb box. Then, the robot opens the box and pulls out a red wire from the bomb, using its manipulator.
\end{enumerate}
Failure conditions include the robot's movement deviating outside the experiment area, the robotic arm moving or breaking the box, and also any sort of hardware failure such as the control system, robot or robotic arm becoming non-functional. 

\subsection{User Profiles}
In total, we have six volunteer users to perform all the above-mentioned four tasks. These users cover different ages, gender, and experience. Half of these users had experience with gamepads. Furthermore, two of them even have robot operating experience. On the other hand, three users have no engineering or gamepad experience.  After basic instruction on controllers, they were directly assigned to tasks without additional practice or training.

\subsection{Gamepad vs IMU Suit Performance Comparison}
\begin{table}
\centering
\caption{Average time to complete each task and its standard deviation in seconds}\label{table:compare}
\vspace{-2mm}
\begin{tabular}{llllll}
             & \multicolumn{2}{c}{Gamepad} & \multicolumn{2}{c}{IMU} & \multicolumn{1}{c}{IMU+VR}  \\ 
\cline{2-6}
             & Avg.      & Stdev.          & Avg.     & Stdev.       & Avg.                        \\ 
\hline
Locomotion   & 14.8      & 4.8             & 32.5     & 13.6         &                             \\
Manipulation & 15        & 5.4             & 12.3     & 5.4          &                             \\
Loco+Mani    & 18.2+18.3 & 12.5            & 34.2+7.8 & 15.4         &                             \\
EOD          & 161.8     & 100.6           & 85.8     & 37.1         & 118                         \\
\hline
\end{tabular}
\end{table}

\begin{figure}
 \centering
 \includegraphics[width=0.92\columnwidth]{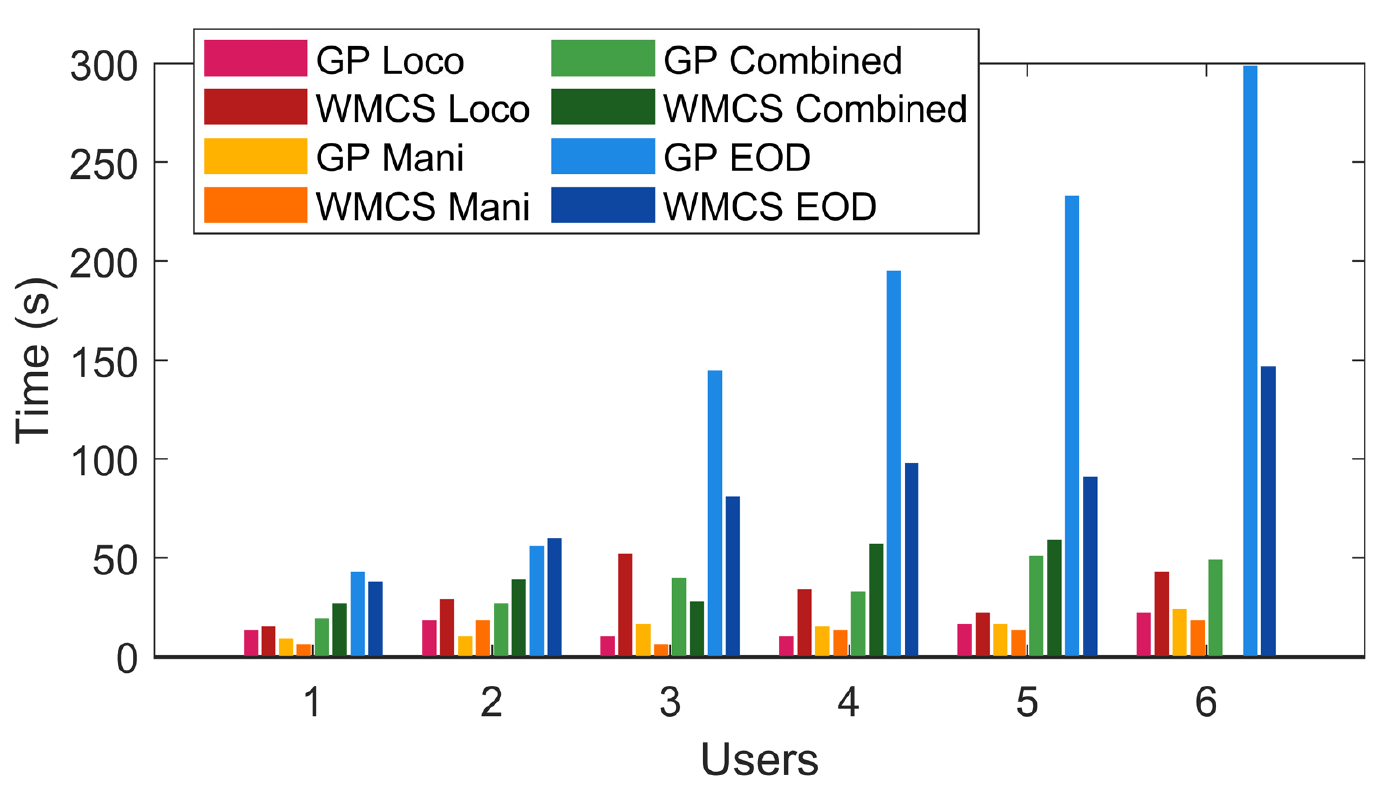}
 \caption{Movement time each user takes to complete tasks with gamepad (GP) and IMU-based wearable motion capture system (WMCS).}
 \label{fig:Big}
\end{figure}

The time taken for all six users to complete each experiment stage is shown in \figref{fig:Big}. User $6$ did not complete the combined test due to tactical difficulty, and all other users completed all planned tasks. \tabref{table:compare} shows the average time of the six users using different controllers to complete each task. From the table, we see that the gamepad has higher performance in the locomotion task and that the wearable motion capture system has the advantage in complex tasks. Also, the user's performance with the wearable motion capture system improves as they practice throughout the experiment.

In the locomotion task, the wearable motion capture system took almost twice as long as the time of gamepad, to finish the task. While the time is similar for both controllers in the manipulation task, with only less than $20\%$ difference. At this point, the gamepad proved to have the advantage of performing better the locomotion tasks. 

\begin{figure*}
 \centering
 \includegraphics[width=\textwidth]{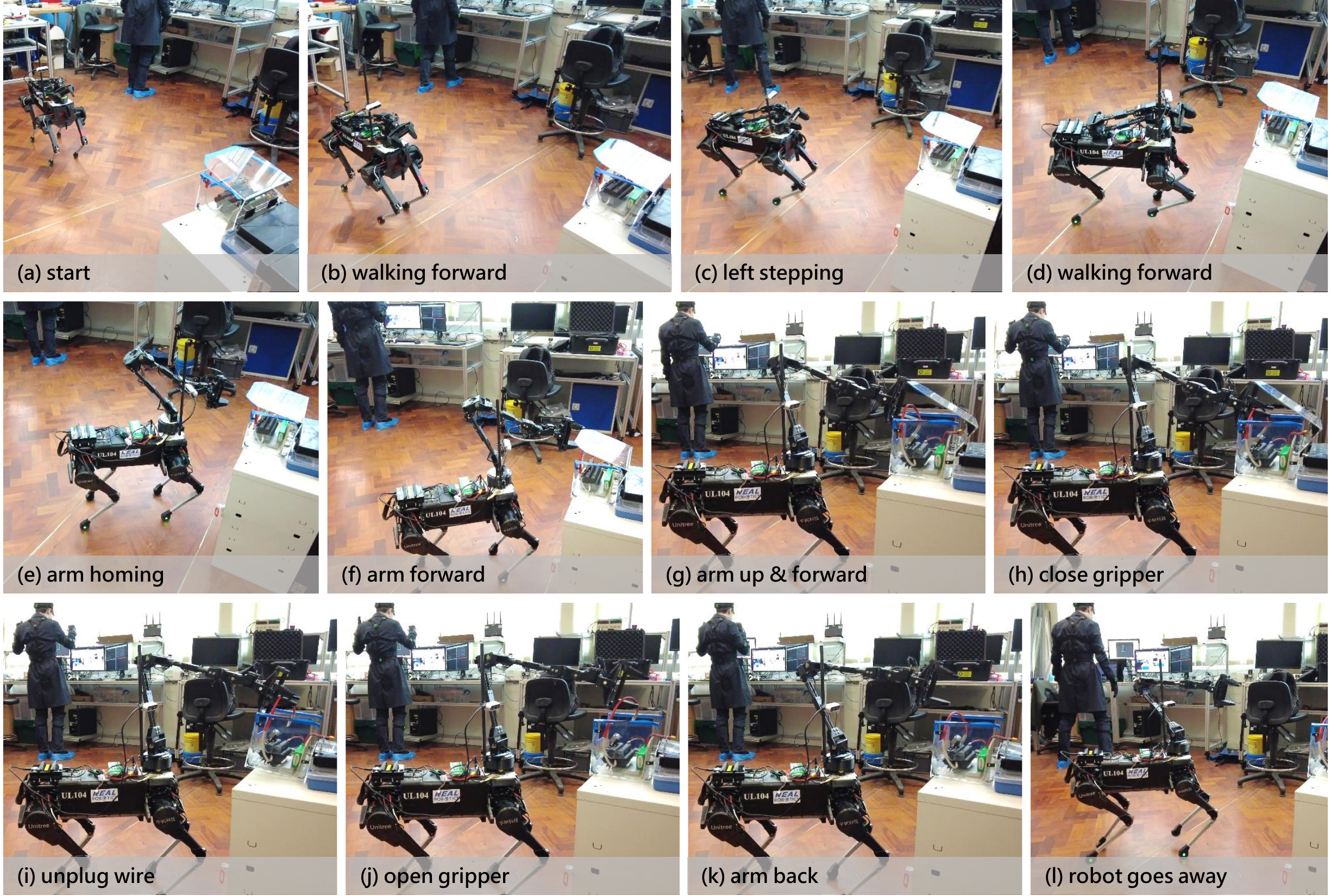}
 \caption{Chronological snapshots of the teleoperator's command with the corresponding teleoperated system's action throughout the live experiment.}
 \label{fig:exp}
\end{figure*}

For the combined locomotion and manipulation task, we see a similar total time between the gamepad and the wearable motion capture system. However, we can break it into two separate parts: the locomotion part and the manipulation part. In this way, we can see that the gamepad has an equal distribution between two parts (locomotion of $18.2$ s and manipulation of $18.3$ s), and the motion capture system experiences difficulty in the locomotion part, but has a significantly better performance in the manipulation part (locomotion of $34.2$ s and the manipulation of $7.8$ s). One reason for such a result is the difference in the Degrees of Freedom, where locomotion is a 2D movement, and manipulation is a 3D movement, which causes manipulation to be more complex than the locomotion movement in this setup. The traditional joysticks on a gamepad usually have $2$DoF control, which makes them suitable for 2D movements. In contrast, the motion capture system captures human arm movement in 3D space, which can easily map to a manipulator's 3D movement. It is also more natural and user-friendly for human teleoperators to use their arm to control the robotic arm in teleoperation. This hardware characteristic makes the wearable motion capture system more suitable for real-world applications. 

One of the example missions is the EOD task, where the robot needs to perform both locomotion and manipulation, however, the manipulation part in this task is more challenging. From \tabref{table:compare}, we can see that the gamepad took almost twice the amount of time to complete the EOD mission in comparison to the wearable motion capture system on average. Some missions similar to this example can be time-sensitive, and timely response can prevent further property damage, personal injury, even fatality. With the advantage of high efficiency, the proposed system that uses wearable motion capture system can be helpful in many real-world applications for teleoperating legged manipulators.

As mentioned earlier, the experiment performs in the order of simpler tasks to more challenging tasks. With this order, users are getting more and more familiar with the system as they practice through the experiment. By examining \tabref{table:compare} for each controller's performance in completing the tasks, we can see that the wearable motion capture system has the tendency to improve performance. In particular, the manipulation part in the combined task has an improvement of over $30\%$ when compared to the earlier manipulation task. However, for the gamepad, there is no evidence of improvement with practice. Furthermore, the performance difference between the two controllers also grows with more practice in favour of the motion capture system. Using the wearable motion capture system, we have one user with only less than two hours further training, who was eventually able to finish the EOD task within $45$ seconds, which almost reduces the time by half. Snapshots of one training trial are shown in Fig.~\ref{fig:exp}. This further proves the practice trend for the motion capture system.

The manipulation tasks we aim at solving with this system requires precision at the level of centimeters. We have successfully experimented with EOD tasks that roughly needed a $5$ cm movement accuracy, which demonstrated the effectiveness of the proposed IMU-based teleoperation.

\subsection{VR Usability}
During the experiment, all users could complete the EOD task with the wearable motion capture system and VR-based integration. From \tabref{table:compare}, they finished the task with an average of $118$ s, which is slower than direct sight control with the wearable motion capture system but still faster than direct sight control with the gamepad. As the camera is located on the side of the arm to have a closer view of the manipulating object, both users think the camera angle is too narrow, especially during locomotion.  However, none of the users had any discomfort during the tests. Most of the users believe the VR is a good choice in a human inaccessible environment.

%%%%%%%%%%%%%%%%%%%%%%%%%%%%%%%%%%%%%%%%%%%%%%%%%%%%%%%%%%%%%%%%%%%%%%%%%%%%%%%%
\section{Conclusions}\label{Sec:concl}
Quadrupedal legged manipulation is an exciting new area of research and development in robotics. In combination with teleoperation, it makes the ideal case of allowing telexistence to replace humans in dangerous or hard tasks. In this paper, we developed such a system and tested it for a typical EOD task. Telexistence is an area of current development in real-world robotics, but also a very challenging one. There are several challenges in providing a reliable teleoperation/VR system on a real robot. 

Some of these challenges were regarding hardware and software integration. The main hardware issue was that legged robots of this scale have a low payload. In our case, the maximum payload of the quadruped robot is only $5$ kg. Thus, only lightweight robot/sensory parts could be installed on its body. We resolved this problem by redesigning the manipulator that was embedded on the robot's back. The main challenge with the software integration was that all submodules installed on three computers (\figref{fig:overview}) had to be synchronised. Some devices were supported in Ubuntu (quadruped, manipulator, RGB-D sensor), while others only in Windows (VR and wearable IMU-based motion capture), and thus they must be bridged synchronously. Our system structure and communication over $5$ GHz Wi-Fi handles this. The wearable IMU-base motion capture system requires calibration before being used, which takes only a few seconds for the teleoperator. Potential drifts due to slight miss-calibration were not preventing the teleoperator from driving the robot accurately, through the use of intuitive relative movement adjustments which eliminate the long term sensor drifting effects. 

Throughout the experiment, the operators have only visual feedback provided either by direct sight or VR. Therefore, safe physical interactions between the robot and the environment heavily rely on the low-level joint torque controller. We are looking at introducing haptic feedback for the operator in future work for a better HRI experience.

There are several directions for advancing the proposed system. Firstly, we are looking into developing code that allows all the devices to work in the same operating system. In this way, the amount of computers that are required will be minimised, and synchronisation issues might decrease. Secondly, we are looking into semi-autonomous teleoperation, where one-to-one mapping between the teleoperator and the robot is avoided. Instead, the robot can use autonomy for some cognitive tasks, while the teleoperator configures the sequences. Last, we aim at using the system for more real-world tasks, such as inspection and monitoring in industrial settings.

In the experiment, we had users with gamepad experience and users without gamepad experience. Despite this difference, they are from the same user group: college students and researchers. This unicity contributes to the limitation of the result, as another group of users with different backgrounds may generate a different result.

%%%%%%%%%%%%%%%%%%%%%%%%%%%%%%%%%%%%%%%%%%%%%%%%%%%%%%%%%%%%%%%%%%%%%%%%%%%%%%%%
\bibliographystyle{IEEEtran}
\bibliography{ram.bib}

\begin{thebibliography}{10}
\providecommand{\url}[1]{#1}
\csname url@samestyle\endcsname
\providecommand{\newblock}{\relax}
\providecommand{\bibinfo}[2]{#2}
\providecommand{\BIBentrySTDinterwordspacing}{\spaceskip=0pt\relax}
\providecommand{\BIBentryALTinterwordstretchfactor}{4}
\providecommand{\BIBentryALTinterwordspacing}{\spaceskip=\fontdimen2\font plus
\BIBentryALTinterwordstretchfactor\fontdimen3\font minus
  \fontdimen4\font\relax}
\providecommand{\BIBforeignlanguage}[2]{{%
\expandafter\ifx\csname l@#1\endcsname\relax
\typeout{** WARNING: IEEEtran.bst: No hyphenation pattern has been}%
\typeout{** loaded for the language `#1'. Using the pattern for}%
\typeout{** the default language instead.}%
\else
\language=\csname l@#1\endcsname
\fi
#2}}
\providecommand{\BIBdecl}{\relax}
\BIBdecl

\bibitem{Mason2012}
M.~T. Mason, ``{Creation Myths: The Beginnings of Robotics Research},''
  \emph{IEEE RAM}, vol.~19, no.~2, pp. 72--77, 2012.

\bibitem{Chestnutt2006}
J.~Chestnutt, P.~Michel, K.~Nishiwaki, J.~Kuffner, and S.~Kagami, ``{An
  Intelligent Joystick for Biped Control},'' in \emph{IEEE International
  Conference on Robotics and Automation}, 2006, pp. 860--865.

\bibitem{Klamt2018}
T.~Klamt \emph{et~al.}, ``{Supervised Autonomous Locomotion and Manipulation
  for Disaster Response with a Centaur-Like Robot},'' in \emph{IEEE/RSJ
  International Conference on Intelligent Robots and Systems}, 2018, pp. 1--8.

\bibitem{Zhang2018}
T.~Zhang, Z.~McCarthy, O.~Jow, D.~Lee, X.~Chen, K.~Goldberg, and P.~Abbeel,
  ``{Deep Imitation Learning for Complex Manipulation Tasks from Virtual
  Reality Teleoperation},'' in \emph{IEEE International Conference on Robotics
  and Automation (ICRA)}, 2018, pp. 5628--5635.

\bibitem{Parnell2018}
E.-J. Rolley-Parnell, D.~Kanoulas \emph{et~al.}, ``Bi-manual articulated robot
  teleoperation using an external {RGB-D} range sensor,'' in \emph{ICARCV},
  2018, pp. 298--304.

\bibitem{anymalwitharm}
C.~D. Bellicoso, K.~Krämer, M.~Stäuble, D.~Sako, F.~Jenelten, M.~Bjelonic,
  and M.~Hutter, ``{ALMA} - articulated locomotion and manipulation for a
  torque-controllable robot,'' in \emph{IEEE International Conference on
  Robotics and Automation}, 2019, pp. 8477--8483.

\bibitem{Klamt2020}
T.~Klamt \emph{et~al.}, ``{Remote Mobile Manipulation with the Centauro Robot:
  Full-body Telepresence and Autonomous Operator Assistance},'' \emph{ArXiv},
  vol. abs/1908.01617, 2020.

\bibitem{Elobaid2018}
\BIBentryALTinterwordspacing
M.~Elobaid, Y.~Hu, J.~Babic, and D.~Pucci, ``{Telexistence and Teleoperation
  for Walking Humanoid Robots},'' \emph{CoRR}, vol. abs/1809.01578, 2018.
  [Online]. Available: \url{http://arxiv.org/abs/1809.01578}
\BIBentrySTDinterwordspacing

\bibitem{Deshpande2018}
N.~Deshpande, J.~Ortiz \emph{et~al.}, ``Next-generation collaborative robotic
  systems for industrial safety and health,'' \emph{WIT Transactions on The
  Built Environment}, vol. 174, pp. 187--200, 2018.

\bibitem{imumocap1}
S.-K. Kim, S.~Hong, and D.~Kim, ``{A Walking Motion Imitation Framework of a
  Humanoid Robot by Human Walking Recognition from {IMU} Motion Data},'' in
  \emph{IEEE-RAS IJHR}, 2009, pp. 343--348.

\bibitem{imumocap2}
S.-O. Shin, D.~Kim, and Y.-H. Seo, ``{Controlling Mobile Robot Using IMU and
  EMG Sensor-Based Gesture Recognition},'' in \emph{International Conference on
  BWCCA}, 2014, pp. 554--557.

\bibitem{imumocap3}
C.~Yang, J.~Chen, and F.~Chen, ``Neural learning enhanced teleoperation control
  of baxter robot using {IMU} based motion capture,'' in \emph{ICAC}, 2016, pp.
  389--394.

\bibitem{imumocap4}
H.~Lv, G.~Yang, H.~Zhou, X.~Huang, H.~Yang, and Z.~Pang, ``Teleoperation of
  collaborative robot for remote dementia care in home environments,''
  \emph{IEEE Journal of Translational Engineering in Health and Medicine},
  vol.~8, pp. 1--10, 2020.

\bibitem{eda}
E.~Dalin, I.~Bergonzani, T.~Anne, S.~Ivaldi, and J.-B. Mouret, ``Whole-body
  teleoperation of the {Talos} humanoid robot: Preliminary results,'' in
  \emph{ICRA 2021 - 5th Workshop on Teleoperation of Dynamic Legged Robots in
  Real Scenarios}, 2021.

\bibitem{Elobaid2019}
M.~Elobaid, Y.~Hu, G.~Romualdi, S.~Dafarra, J.~Babic, and D.~Pucci,
  ``Telexistence and teleoperation for walking humanoid robots,'' in
  \emph{Intelligent Systems and Applications}, vol. 1038, 2019, pp. 8477--8483.

\bibitem{Tachi2013}
S.~Tachi, K.~Mlnamlzawa, M.~Furukawa, and C.~L. Fernando, ``{Haptic Media
  Construction and Utilization of Human-Harmonized ``Tangible'' Information
  Environment},'' in \emph{International Conference on Artificial Reality and
  Telexistence}, 2013, pp. 145--150.

\bibitem{sutherland1984}
I.~E. Sutherland and M.~K. Ullner, ``{Footprints in the Asphalt},''
  \emph{International Journal of Robotics Research}, vol.~3, no.~2, pp. 29--36,
  1984.

\bibitem{zhou2017jbe}
C.~Zhou, X.~Wang, Z.~Li, and N.~Tsagarakis, ``Overview of gait synthesis for
  the humanoid {COMAN},'' \emph{Journal of Bionic Engineering}, vol.~14, no.~1,
  pp. 15--25, 2017.

\bibitem{You2016robio}
Y.~You, S.~Xin, C.~Zhou, and N.~Tsagarakis, ``{Straight Leg Walking Strategy
  for Torque-controlled Humanoid Robots},'' in \emph{IEEE International
  Conference on Robotics and Biomimetics}, 2016, pp. 2014--2019.

\bibitem{Liu2020}
J.~Liu, P.~Balatti \emph{et~al.}, ``Garbage collection and sorting with a
  mobile manipulator using deep learning and whole-body control,'' in
  \emph{IEEE-RAS International Conference on Humanoid Robots}, 2020.

\end{thebibliography}

\end{document}